\documentclass[letterpaper, 10 pt, conference]{ieeeconf}  
\IEEEoverridecommandlockouts
\usepackage[noadjust]{cite}
\usepackage{amssymb}
\usepackage{amsmath}
\usepackage{graphics}
\usepackage{epsfig}
\usepackage{breqn}
\usepackage[T1]{fontenc}
\usepackage{booktabs,multirow,array}
\usepackage[pagebackref=true,colorlinks,citecolor=green]{hyperref}
\hypersetup{%
  colorlinks = True,
  linkcolor  = blue
}

\usepackage{xcolor,colortbl, fancybox}
\definecolor{Gray}{gray}{0.85}
\definecolor{LightCyan}{rgb}{0.88,1,1}
\usepackage[justification=centering]{caption}
\let\labelindent\relax
\usepackage[inline]{enumitem}  
\usepackage{pgfplots}
\usepackage[normalem]{ ulem }
\usepackage{soul}

\newcommand{\R}{\mathbb{R}}

\title{\LARGE \bf
Analysis over vision-based models for pedestrian action anticipation
}

\author{Lina Achaji$^{1,2}$, Julien Moreau$^{1}$, François Aioun$^{1}$, and Franois Charpillet$^{2}$
\thanks{$^{1}$Stellantis, France
{\tt\small lina.achaji@stellantis.com}}
\thanks{$^{2}$Université de Lorraine, CNRS, Inria, LORIA, Nancy 54000, France}
}

\begin{document}

\maketitle
\thispagestyle{empty}
\pagestyle{empty}

\begin{abstract}

Anticipating human actions in front of autonomous vehicles is a challenging task. Several papers have recently proposed model architectures to address this problem by combining multiple input features to predict pedestrian crossing actions. This paper focuses specifically on using images of the pedestrian's context as an input feature. We present several spatio-temporal model architectures that utilize standard CNN and Transformer modules to serve as a backbone for pedestrian anticipation. However, the objective of this paper is not to surpass state-of-the-art benchmarks but rather to analyze the positive and negative predictions of these models. Therefore, we provide insights on the explainability of vision-based Transformer models in the context of pedestrian action prediction. We will highlight cases where the model can achieve correct quantitative results but falls short in providing human-like explanations qualitatively, emphasizing the importance of investing in explainability for pedestrian action anticipation problems.
 
\end{abstract}


\section{Introduction}
The European Commission's 2019 road safety statistics \cite{eustat} indicate that a significant proportion of road fatalities occur in urban areas, with pedestrians being the most vulnerable group. Predicting pedestrian actions can help reduce the percentage of pedestrian fatalities in urban areas by enabling autonomous vehicles and human drivers to anticipate and avoid potential accidents.

\subsection{Motivation and related work}
Over the past few years, there have been numerous studies proposing the use of ego-vehicle view camera sensor data fed to deep learning models to predict future human crossing actions. Pedestrian behavior can be influenced by various factors, including past states, social norms, and environmental factors. To address this, researchers in \cite{rasouli2020pedestrian}, \cite{kotseruba2021benchmark}, \cite{yang2022predicting} have focused on creating model architectures that can combine multiple data features such as bounding boxes, pose skeletons, and raw images to predict pedestrian actions. Recently, the authors in \cite{achaji} and \cite{capformer} leveraged Transformer networks for the action anticipation task. The architecture in \cite{capformer} is composed of various branches which fuse video and kinematic data. However, all of these models have not qualitatively analyzed each of the used features.  
In this paper, we will focus our study on one of these features. Namely, our work will address the usage of raw images as implicit features to the predictive model. In fact, using images as input to the model can be advantageous since they do not require labeling. Additionally, using a temporal sequence of raw images can serve as a spatio-temporal feature to the model. In this paper, we will compare the use of image features as input to multiple models, including spatio-temporal CNN-based models, spatio-temporal Transformer-based models, and combinations of CNN and Transformer modules. We will analyze, based on the model’s output predictions, the operational domain for each model and determine where raw images can successfully predict pedestrian actions and where they fail.

Explainable deep learning methods have recently been effective in visualizing the internal decision-making processes of CNN and Transformer models \cite{selvaraju2017grad}, \cite{abnar2020quantifying}, \cite{chefer2021transformer}. However, these techniques have not been applied to the spatio-temporal domain in general and specifically not to the prediction of pedestrian actions. In this paper, we will utilize a spatio-temporal Transformer-based model to provide explanations for our findings. Our analysis highlights that relying solely on quantitative results can be risky in real-world scenarios without further testing. Specifically, our results demonstrate that vision-based models may produce favorable quantitative results but fail to provide human-like explanations in certain situations.

\subsection{Contributions}

\begin{enumerate}
    \item Our research aims to compare different deep learning models that are designed for predicting pedestrian actions using spatio-temporal data. Specifically, we focus on models that take image pixels as input and compare CNN-based, Transformer-based, and hybrid models that combine both architectures.
    \item We provide a thorough analysis of each proposed model's strengths and weaknesses in predicting pedestrian actions. We examine two types of input images: those that emphasize the pedestrian and those that focus on the pedestrian's surroundings. Our quantitative results demonstrate that image-based models perform well on the test dataset. However, we also highlight certain qualitative limitations of Transformer-based models, which raises concerns about their explainability in predicting pedestrian actions.
\end{enumerate} 

\section{Problem statement}

This section assesses various computer vision architectures for predicting pedestrian actions using raw images as input. The action anticipation block receives a sequence $X = \{x_1, x_2, \dots, x_k \}$ of $k$ RGB images during the observation interval $T_{\text{obs}}$, where $x_i \in \R^{w \times h \times 3}$ is an image frame at time $t=i$ within $X$. The objective is to calculate the probability of the pedestrian crossing the street as a binary classification problem $P(Y | X)$, conditioning on the entire observation history. If the pedestrian crosses the street, the event time $A$ represents the start time of the crossing. If not, $A$ represents the time of the last observable frame of the pedestrian. The duration between the last observed frame and the critical time $A$ is referred to as the Time-To-Event (TTE).

We employ two different techniques to crop the images around the pedestrian bounding box of interest, which is defined by the $x$ and $y$ coordinates of the upper-left corner, as well as the corresponding coordinates of the lower-right corner of the rectangular bounding box.
\begin{enumerate}
    \item The first method uses a static width $w_c$ and height $h_c$ crop, which is centered on the bounding box's center coordinate. The values of $w_c$ and $h_c$ are greater than the bounding box's width $b_w$ and height $b_h$, respectively. By using this method, a portion of the environment surrounding the pedestrian is included in the input passed to the model.
    \item The second method employs dynamic width and height with respect to each of the pedestrians, where $w_c$ = $b_w$ and $h_c$ = $b_h$. This method enables the model to access only the information about the pedestrian without relying on the environment. However, since the pretrained models require a fixed input shape, we add padding around the dynamic crop to maintain a consistent size across all samples.
\end{enumerate}

\section{Proposed models}
This section aims to provide an overview on the novel architectures that we have devised for action anticipation.

\subsection{I3D augmented by Transformers -- I3D-Trans}

\begin{figure}[b]
\includegraphics[scale=0.3]{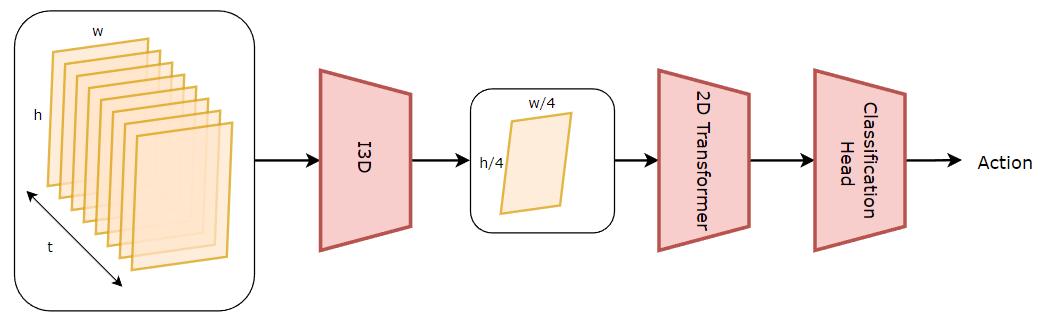}
\centering
\caption{Overview of the I3D-Trans architecture.}
\label{fig:I3D-Trans}
\end{figure}

We propose the I3D-Trans model (Fig. \ref{fig:I3D-Trans}) which is composed of multiple stages. The primary objective of the initial phase of the I3D-Trans model is to condense the temporal dimension of a sequence of 3D images into a single image. Afterward, the resulting image undergoes encoding of spatial dependencies by a 2D Transformer. Finally, the output of the Transformer is fed into a classification head that predicts pedestrian actions.

\subsubsection{Temporal modeling}

The I3D-Trans model utilizes the I3D architecture \cite{carreira2017quo} as its backbone and processes the sequence of 3D images in temporal order to reduce the temporal dimension. Initially proposed for video processing, the I3D backbone is a 3D CNN architecture consisting of inflated CNN branches that incorporate pretrained inception \cite{szegedy2015going} modules. A significant contribution of the I3D model is its training on the Kinetics human action dataset \cite{kay2017kinetics}. In our study, we use a pretrained I3D model and fine-tune the backbone throughout the training process. However, in the encoding process, we don't use the pretrained classification-head used by the I3D model. As a consequence, the resulting image has a D-dimensional representation with a narrower latent width, $w_{l}$, and latent height, $h_{l}$, compared to the input.

\subsubsection{Spatial modeling}


The d-dimensional encoded image undergoes processing by a Transformer that incorporates a custom sequential attention network to calculate dependencies among the rows and columns of the image. This Transformer module is trained from scratch as apposed to the pretrained I3D backbone. Initially, the original image $x \in \R^{w_{l} \times h_{l} \times d}$ passes through an attention layer that computes dependencies among pixels positioned at the same row and varying column. The second attention layer utilizes the output of the first layer to compute dependencies among all the pixels located at the same column. The Multi-head attention mechanism introduced in \cite{vaswani2017attention} is employed in each of the attention layers.

The described Transformer network utilizes positional encoding to maintain the spatial location of each pixel during the encoding process. The positional encoding formula, the point-wise feed-forward neural networks (P-FFN), and the layer normalization layers are identical to the one used for temporal encoding in \cite{vaswani2017attention}. The overall module is governed by the following equations: 

$$ x = x W_e + b_e $$
$$ x_{att} = \text{MHA}(QKV=x)  $$
$$ x_{att}^T = \text{MHA}(QKV=\text{permute}(x_{att})) $$
$$ x = \text{LayerNorm}(x + \text{permute}(x_{att}^T)) $$
$$ x = \text{LayerNorm}( x + \text{P-FFN}(x)) $$

\subsubsection{Classification head}
The Transformer module generates an image of size $(w_l, h_l, d)$ as its output. In order to obtain a feature vector of size $d$, we perform global average pooling on each dimension of the image. This resulting vector is then fed into a linear and sigmoid layer for binary classification of the action.

\subsection{Inception augmented by Transformers -- Inception-Trans}
In contrast to the I3D-Trans model, the Inception-Trans architecture computes the spatial correlations of each image separately in the initial stage. Later, a temporal Transformer is employed to capture the temporal dependencies among the latent vectors generated by the inception module.

\subsubsection{Spatial modeling}
For every image $x$ with dimensions of $(w, h, 3)$, a convolutional module will be utilized to condense its spatial features into a single latent vector of size $d$. This convolutional module can be any contemporary CNN module, such as Inception \cite{szegedy2015going} or ResNet \cite{he2016deep}, that has been pre-trained on an extensive set of images, such as the ImageNet dataset \cite{deng2009imagenet}. The output of the spatial module $h$ is a vector of length $t$, which is equivalent to the initial sequence length and has a dimensionality of $d$.

\begin{figure}[t]
\includegraphics[scale=0.35]{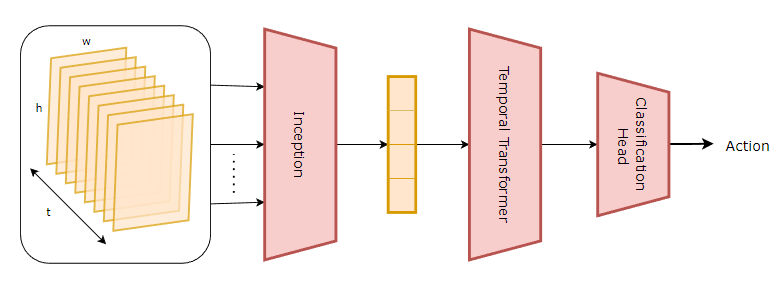}
\centering
\caption{Overview of the Inception-Trans architecture.}
\label{fig:Inception-Trans}
\end{figure}

\subsubsection{Temporal modeling}
To model the temporal aspect, we use a temporal Transformer, which takes the sequence $h$ with a length of $t$ as input and utilizes self-attention to compute the temporal dependencies among the encoded images. The temporal Transformer uses the identical architecture as the one used in  \cite{vaswani2017attention}.

\subsubsection{Classification head}
The temporal Transformer produces an output with the same shape as its input, which is a two-dimensional tensor with dimensions $(t, d)$. To perform binary classification, the classification head applies a mean pooling operation over the sequence length first, followed by passing the result through a dense layer with a sigmoid non-linearity to reduce the dimension from $d$ to 1.

\subsection{Video vision Transformer - ViVIT}
Recently, the ViVIT model \cite{arnab2021vivit} was proposed as an extension of ViT \cite{dosovitskiy2020image}, which was originally designed for image classification, for video classification tasks. ViVIT employs a sequence of pure Transformer layers that incorporate both spatial and channel attention to process images, enabling the model to attend to relevant spatial locations and channels in the image. The video vision Transformer is trained on the Kinetics dataset \cite{kay2017kinetics}. 

\section{Experiments}
In this section, we assess the performance of several models with an observation duration of 16 frames (equal to 0.5 seconds) and a Time-To-Event (TTE) ranging from 30 to 60 frames (equal to 1 to 2 seconds). However, we used a stride of 2 steps on the observation scene for each input sequence, which resulted in a total of 8 frames during the 0.5 seconds of observation. For the static crop setting, we cropped the images with a size of $(600,600)$ around the center of the pedestrian bounding box and resized them to match the input size of each pre-trained model (e.g., inception, I3D, or ViVIT). On the other hand, for the dynamic crop, we enlarged the crop size by $5\%$ of the bounding box height. To balance the dataset for the training set, we applied a flip-based augmentation on the images, followed by an under-sampling technique on the majority class. During training, we used a batch size of 8 and employed the Adam optimizer with an initial learning rate of $10^{-4}$. The models were trained for a total of 20 epochs. Our experimental approach will adhere to the action prediction benchmark \cite{kotseruba2021benchmark}, utilizing the default dataset split configuration where set03 is designated as the testing split. Moreover, we will conduct experiments with random split settings, allocating 0.7 ratio for the training set and 0.2 ratio for the testing set.

\subsection{Results}

\begin{table}[b]
\caption{Model metrics results when using static versus dynamic crop on the default split settings. Accuracy (ACC), Area under curve (AUC), and F1-score.} 
\begin{center}
\begin{tabular}{c|c|c|c} 
 \hline
 \multicolumn{4}{c}{Static} \\
 \hline
  \rowcolor{Gray}
 Model & ACC & AUC & F1-score \\ [0.5ex] 
 \hline\hline
I3D & 0.79 & 0.69 & 0.8 \\
 \hline
I3D-Trans & 0.78 & 0.68 & 0.79 \\
 \hline
Inception-Trans & 0.77 & 0.61 & 0.73 \\
 \hline
ViVIT & 0.8 & 0.66 & 0.77 \\
 \hline
  \hline
 \multicolumn{4}{c}{Dynamic} \\
 \hline
  \rowcolor{Gray}
 Model & ACC & AUC & F1-score \\ [0.5ex] 
 \hline\hline
I3D & 0.81 & 0.72 & 0.82 \\
 \hline
I3D-Trans & 0.80 & 0.70 & 0.81 \\
 \hline
Inception-Trans & 0.73 & 0.63 & 0.76 \\
 \hline
ViVIT & 0.8 & 0.69 & 0.8 \\
 \hline
\end{tabular}
\end{center}
\label{image_results}
\end{table}

\begin{figure}[t]
\centering
\begin{tabular}{cc}
{\includegraphics[scale=0.22]{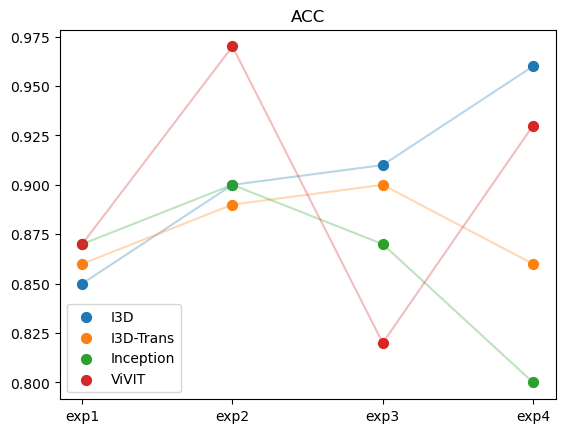}} &
{\includegraphics[scale=0.22]{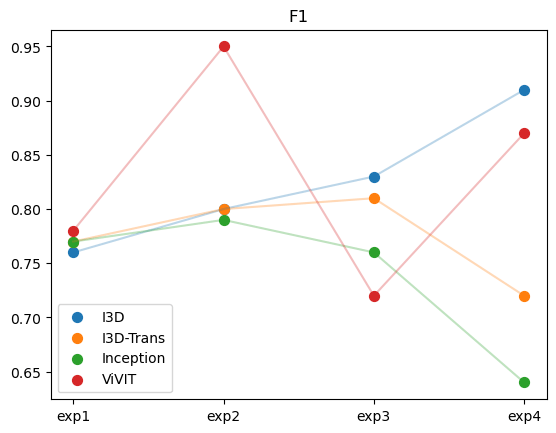}} \\
{\includegraphics[scale=0.22]{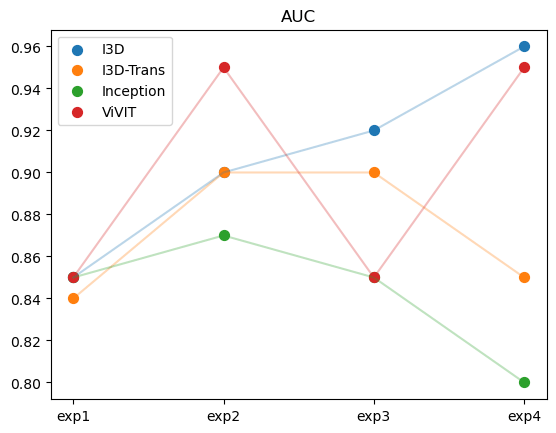}} \\
\end{tabular}
\caption{Accuracy, F1-score, and AUC evaluation for all models on various random splits.} 
\label{fig:random_metrics}
\end{figure}

Initially, we assessed our models using the default training/testing split, and by altering the image crop type. The results, as depicted in Table \ref{image_results}, indicate that the dynamic crop led to better performance across all models than the static crop. Furthermore, in both the static and dynamic settings, the I3D model outperformed the pure Transformer model (ViVIT) and the hybrid CNN-Transformer models consistently. We also evaluated the models using different random training/test splits with the PIE dataset. As presented in Fig. \ref{fig:random_metrics}, the I3D model was not consistently the best performing model, as the rank and performance metric (Accuracy, AUC, and F1-score) varied significantly between experiments. Nonetheless, we observed that the ViVIT model and the I3D models were the top-performing models. This observation could be due to two factors. Firstly, the I3D and ViVIT models are not hybrid models, and attaching a Transformer head to a CNN network could impair overall performance. However, this cannot be a definitive reason since these models are also pre-trained. In contrast to I3D-Trans or Inception-Trans, the Transformer head was trained from scratch on the PIE dataset. To make a fair comparison between the four models, it would be necessary to pre-train the Inception-Trans and I3D-Trans models on a large corpus dataset such as Kinetics and then fine-tune the models on the PIE dataset. However, this was not feasible due to limited training resources.

\begin{figure}[b]
\includegraphics[scale=0.5]{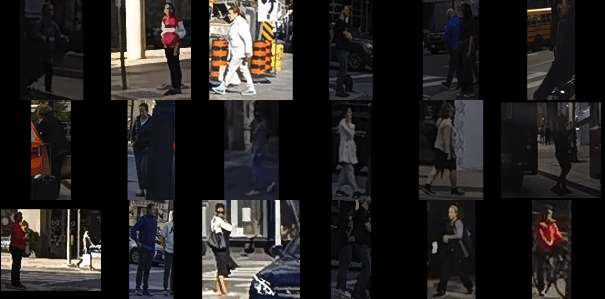}
\centering
\caption{Instances of scenarios where all models predictions are incorrect.}
\label{fig:all_models_images}
\end{figure}

In this section, we will be conducting an analysis of the models' overall performance. When taking the predictions for all of the four models on the dynamic crop settings, only 7\% of the predictions were false for all of the models at the same time, and out of these, 75\% corresponded to non-crossing behavior. The all models' low rate of false predictions implies that developing an ensemble learning approach may prove effective in this scenario. 

Figure \ref{fig:all_models_images} illustrates situations in which none of the models accurately predicted the pedestrian's action. These cases are likely due to several factors, including blurry images or partial/total occlusion of the pedestrian. Additionally, many of the pedestrians in these scenarios are not directly interacting with the ego-vehicle, making it challenging for the models to make accurate predictions without access to the environmental context within the dynamic crop.

\begin{figure}[t]
\includegraphics[scale=0.15]{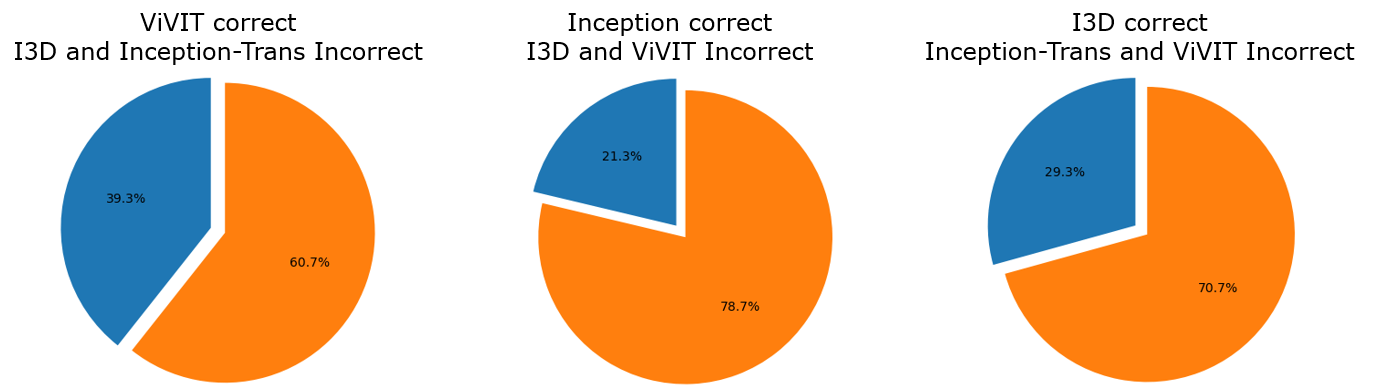}
\centering
\caption{The ratio of predictive accuracy for each model when all other models are incorrect.}
\label{fig:all_models_correct}
\end{figure}

In addition, we computed the percentage of instances in which each model made accurate predictions while all other models failed to do so. Fig. \ref{fig:all_models_correct} illustrates that the ViVIT model achieves the highest accuracy ratio (39.3\%) in situations where both the I3D model and Inception-Trans models have failed. This result indicates that the ViVIT model is more effective in predicting challenging scenarios with greater accuracy. To avoid redundancy, we excluded the I3D-Trans model from our analysis as it shares similar features with the I3D model.

\begin{figure}[t]
\centering
\begin{tabular}{cc}
{\includegraphics[scale=0.2]{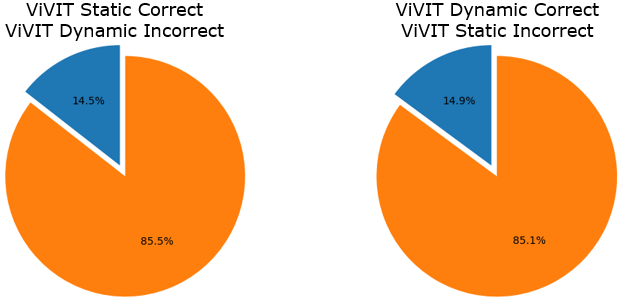}} &
{\includegraphics[scale=0.2]{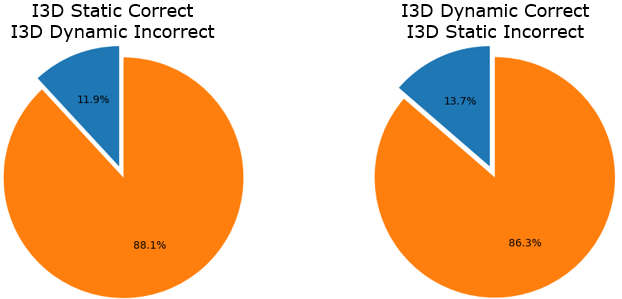}} 
\end{tabular}
\caption{Percentage of accurate predictions for I3D and ViVIT mode in one image cropping mode while being inaccurate in the other mode.} 
\label{fig:vivit_dynamic_static}
\end{figure}

In order to thoroughly examine the performance of models when utilizing static and dynamic crops, we computed the percentage of instances in which each model made correct predictions under dynamic mode while simultaneously making incorrect predictions under static mode, and vice versa. The potential for complementary predictions can be observed in Fig. \ref{fig:vivit_dynamic_static} when using the models in either dynamic or static mode. The models exhibit relatively high performance in making correct predictions when used exclusively in one mode. Notably, the Inception dynamic model can account for a significant portion (approximately 22\%) of errors in dynamic predictions. A future research question could address the optimal approach to jointly predict actions based on dynamic and static crop model branches.

\begin{figure}[b]
\includegraphics[scale=0.5]{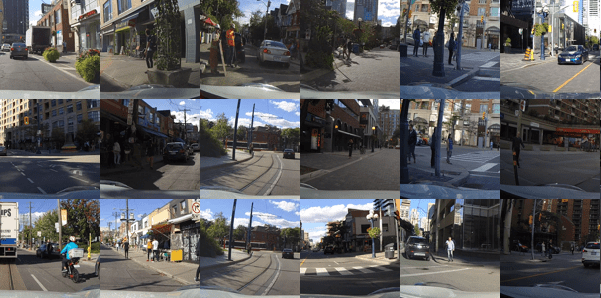}
\centering
\caption{Instances of scenarios where the ViVIT model was correct in static mode while being incorrect in dynamic mode.}
\label{fig:vivit_static}
\end{figure}

Figure \ref{fig:vivit_static} interestingly shows instances where the ViVIT model was able to correctly identify static crops but failed to accurately predict dynamic crops. These cases clearly highlight the challenging nature of the environment, particularly in scenarios such as crosswalks or areas with frequent turning movements.

\section{Can we reach explainability?}

CNN-based models and Transformers are both effective in predicting image-based actions, but Transformers offer an advantage in interpretability due to their attention mechanism. This mechanism enables visualization of the attention weights assigned to each input token, providing insights into the model's decision-making process. In this section, we will investigate the predictions made by the ViVIT model using spatial and temporal attention map visualization. To accomplish this, we will utilize the attention rollout method outlined in \cite{abnar2020quantifying}. This method takes into account both the attention maps and residual connections in the Transformer model to determine how information flows from the input layer to the embeddings in higher layers. For each of the following figures we will show the original temporal sequence of images in the first row, followed by the model attention heatmaps for this sequence in the second row.

\begin{figure}[h]
\centering
\begin{tabular}{c}
{\includegraphics[scale=0.125]{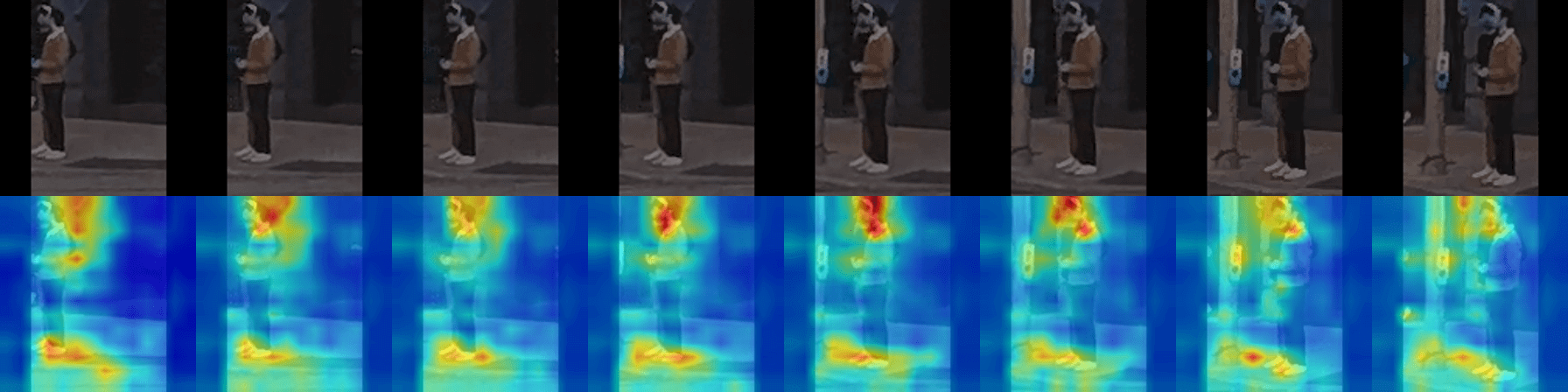}} \\
{\includegraphics[scale=0.125]{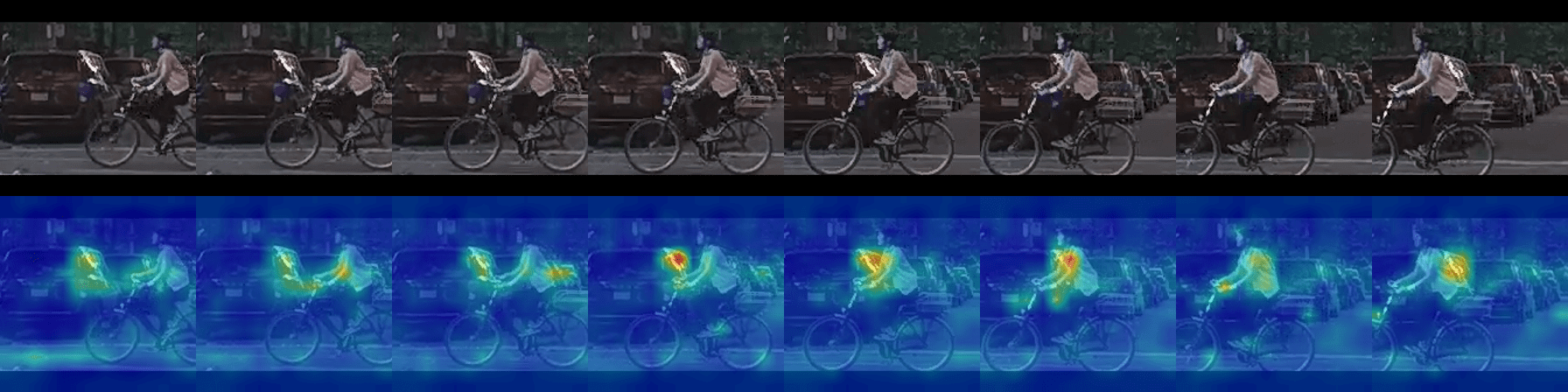}} \\
{\includegraphics[scale=0.125]{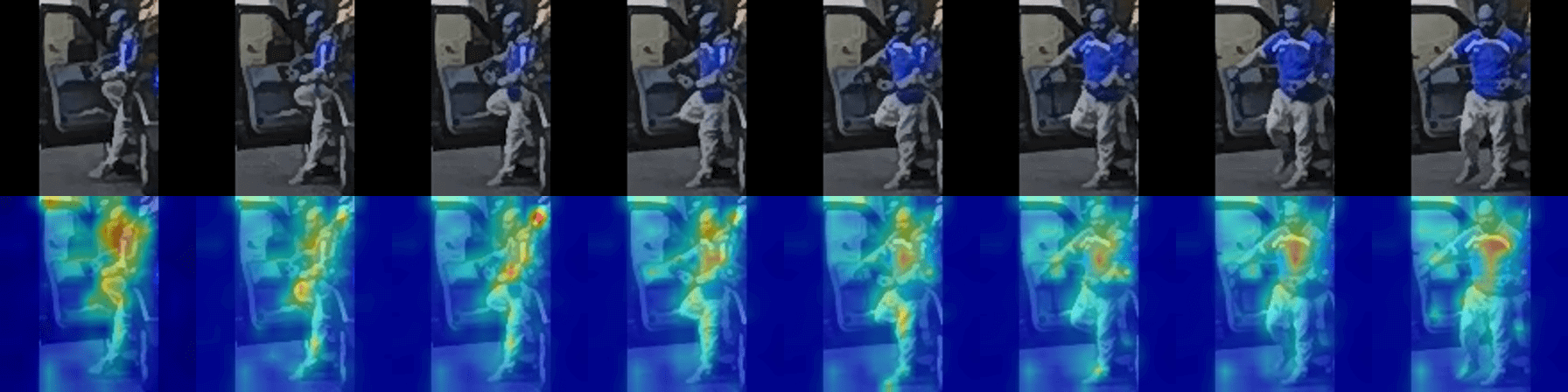}} 
\end{tabular}
\caption{Visualization of spatial attention maps across multiple time steps while utilizing dynamic cropping and achieving accurate predictions.} 
\label{fig:dynamic_correct}
\end{figure}

\begin{figure}[h]
\centering
\begin{tabular}{c}
{\includegraphics[scale=0.125]{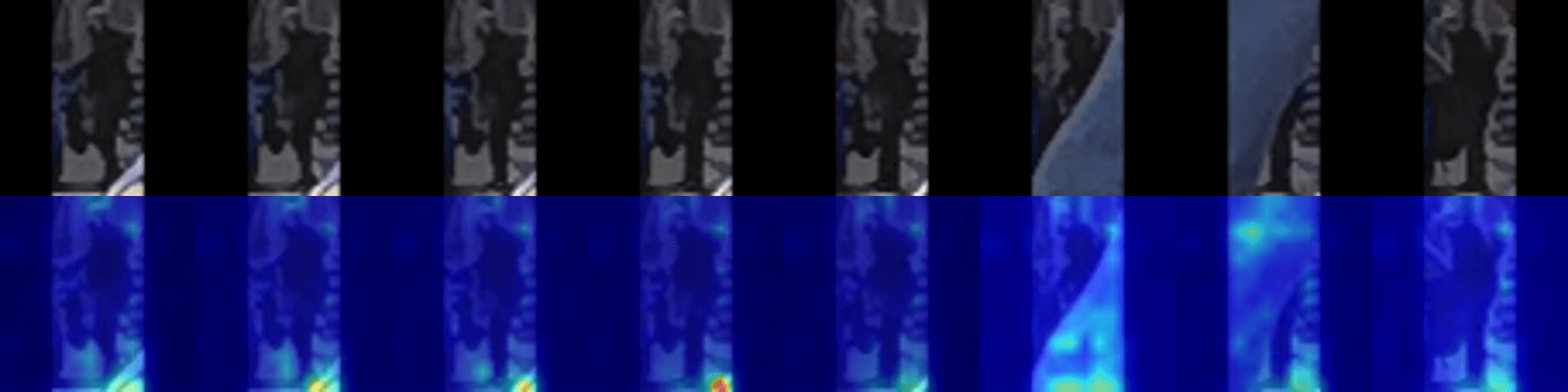}} \\
{\includegraphics[scale=0.125]{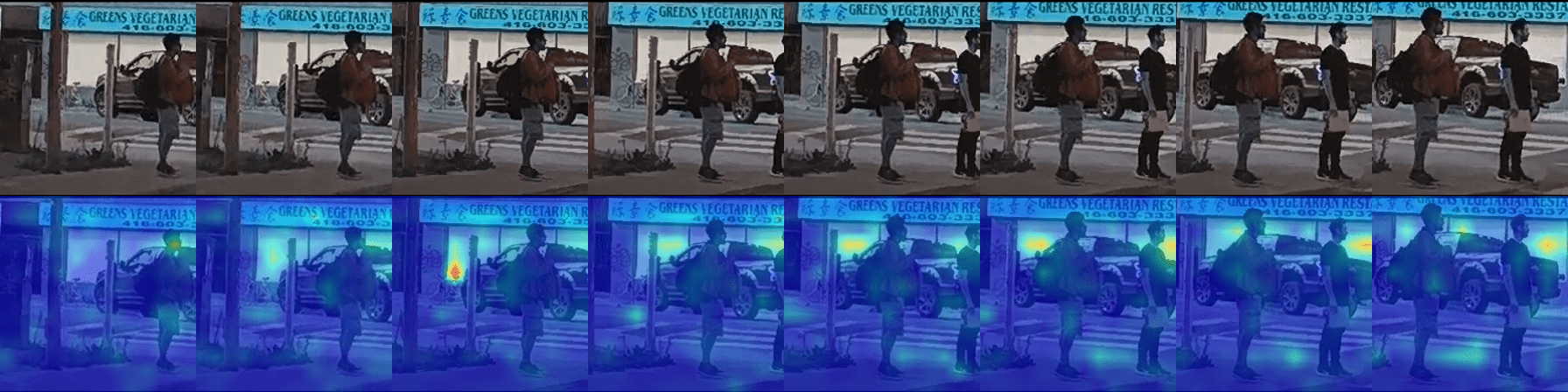}} 
\end{tabular}
\caption{Visualization of spatial attention maps across multiple time steps while utilizing dynamic cropping and achieving incorrect predictions.} 
\label{fig:dynamic_incorrect}
\end{figure}

\begin{figure}[h]
\centering
\begin{tabular}{c}
{\includegraphics[scale=0.125]{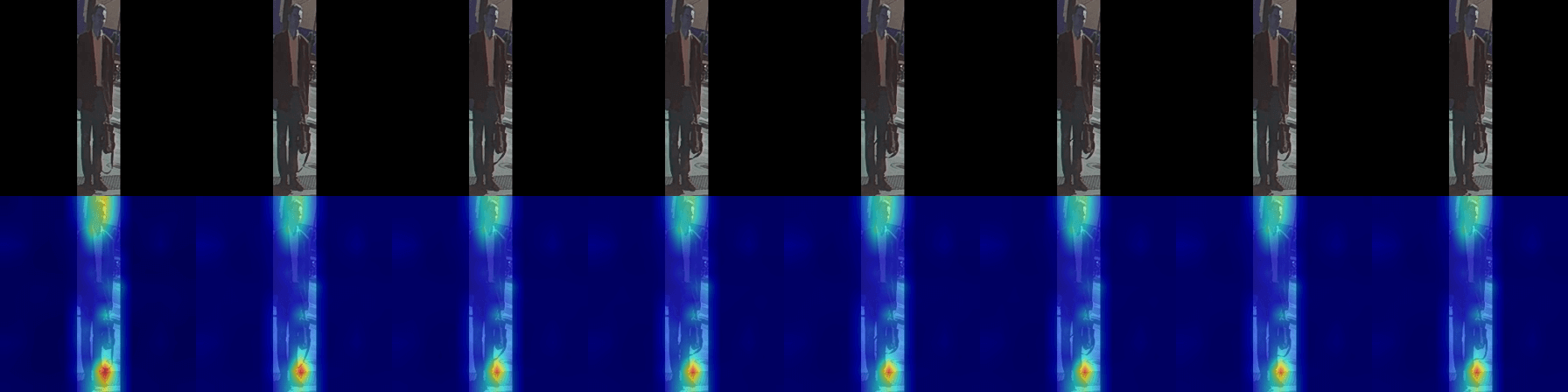}} \\
{\includegraphics[scale=0.125]{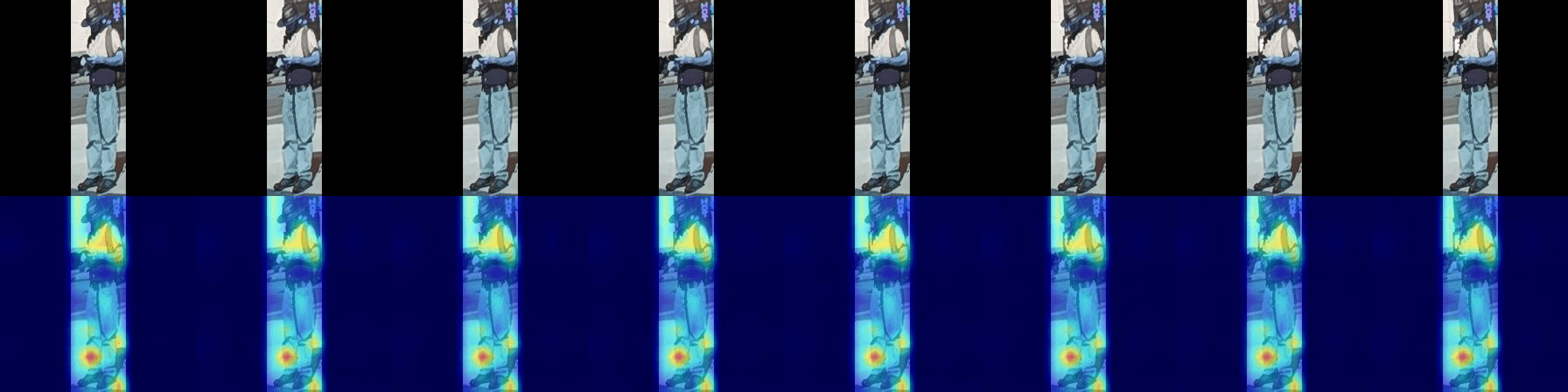}} 
\end{tabular}
\caption{Visualization of spatial attention maps across multiple time steps while utilizing dynamic cropping and achieving incorrect predictions.} 
\label{fig:dynamic_incorrect2}
\end{figure}

To begin our examination, we will first visualize the attention maps produced when using the dynamic crop settings. As illustrated in Fig. \ref{fig:dynamic_correct}, the attention maps reveal that the model has focused on the pedestrians' head and legs in the images when making accurate predictions. These visualizations are encouraging as they align with what humans typically use to predict pedestrian actions or activities. Additionally, the attention's localization density has shifted over time to track the pedestrian's movements, as demonstrated in the last image of Fig. \ref{fig:dynamic_correct}. This confirms that the model has the ability to recognize actions over time, making it suitable for action recognition tasks that require tracking of motion and gestures. We also examined instances where the model produced inaccurate predictions. As depicted in Fig. \ref{fig:dynamic_incorrect}, these examples demonstrate situations where the model failed to anticipate pedestrian actions. The visualizations reveal that the model encountered challenges in identifying the relevant pedestrian in the scene and focused on irrelevant areas, making it unlikely to predict the correct action. Despite successfully localizing the pedestrian in some cases, the model was still unable to predict the action accurately (as seen in Fig. \ref{fig:dynamic_incorrect2}). This could be due to various factors, including the complexity of the scene, which rendered the dynamic crop option unsuitable. Moreover, even if the model correctly localized the pedestrian, it might still have difficulty identifying their gesture or action.

\begin{figure}[h]
\centering
\begin{tabular}{c}
{\includegraphics[scale=0.125]{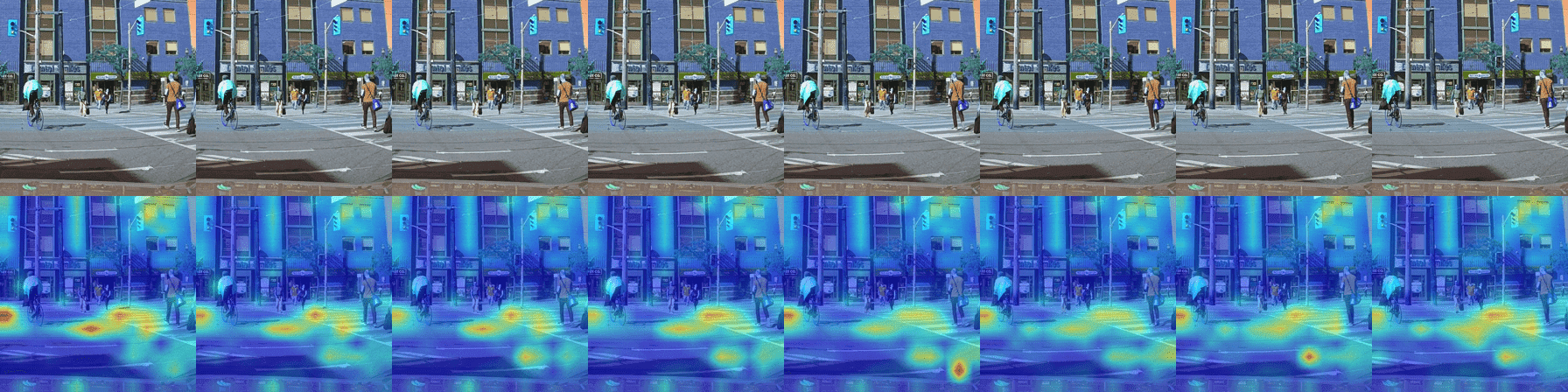}} \\
{\includegraphics[scale=0.125]{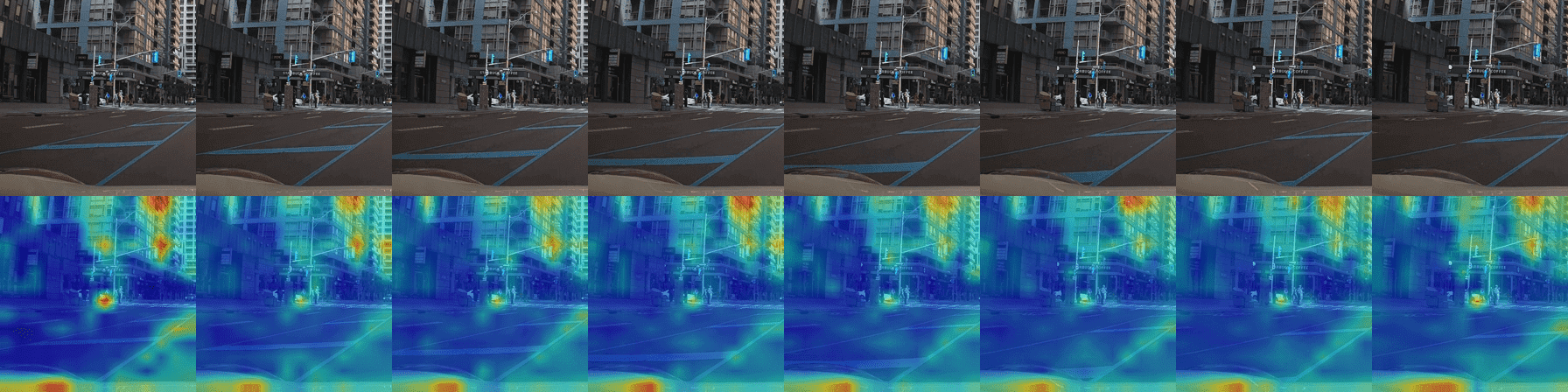}} \\
{\includegraphics[scale=0.125]{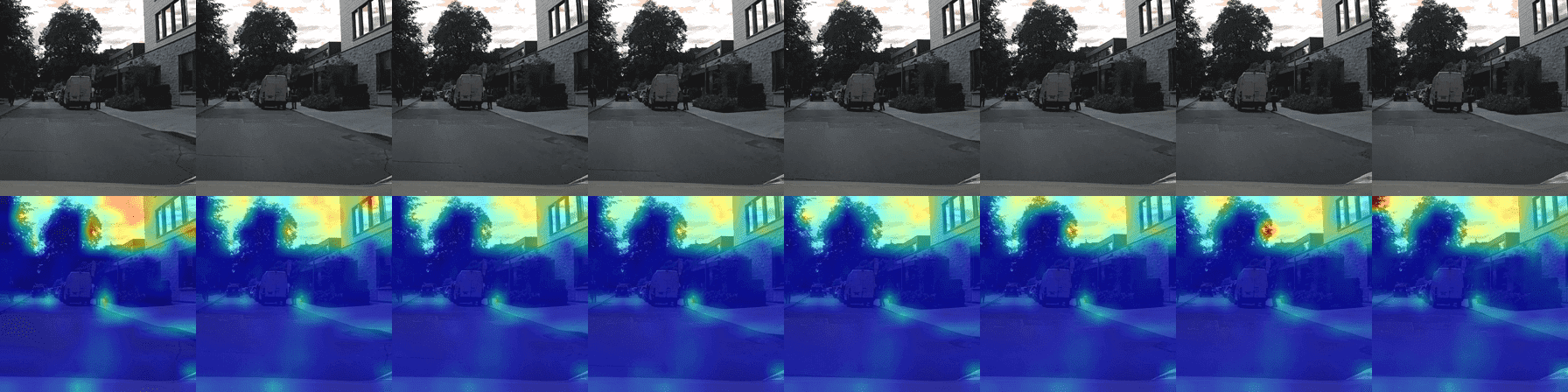}} \\
\end{tabular}
\caption{Visualization of spatial attention maps across multiple time steps while utilizing static cropping and achieving correct predictions.} 
\label{fig:static_correct}
\end{figure}

Following that, we examined the attention maps generated by the model when trained and tested on the static crop settings. Fig. \ref{fig:static_correct} illustrates instances where the model made correct predictions. Interestingly, we observed that in some cases the model was able to successfully segment the crosswalk or intersection lines without any supervision on these features during training. However, the static crop setting has significant limitations when compared to the dynamic crop setting. Firstly, the model frequently failed to accurately localize the relevant pedestrian in almost all cases. While it may be understandable for the model to fail in localizing the pedestrian in instances where the prediction was incorrect, it was puzzling to observe that the model achieved correct predictions despite fixating on the road and ignoring the pedestrian. This suggests that the model may rely solely on environmental cues to predict actions, rather than the pedestrian's characteristics. If this is the case, then the PIE dataset may be biased towards environmental contexts, meaning that the presence of a crosswalk may be sufficient for the pedestrian to cross the street without any specific features related to the pedestrian. Furthermore, using the static crop setting raises more concerns when we consider that the model often fixated on irrelevant features such as the sky, buildings, or parts of the ego-vehicle in order to make a prediction about whether the pedestrian would cross the street. It is worth noting that the PIE dataset was captured in clear weather conditions, and variations in sky color or building shape and color should not impact the model's decision-making. Additionally, we observed that the model was biased towards brighter colors in the images, which may be attributed to the fact that the PIE dataset is relatively small, and the model may not have learned efficient features during the fine-tuning phase. Moreover, unlike the dynamic crop mode, which focuses on the pedestrian and captures their action, the static crop setting did not provide the model with similar types of images that were present in the original kinetics dataset used to pre-train the ViVIT model. Hence, we suggest that a model like I3D or ViVIT should be trained from scratch on a large-scale dataset specifically designed for autonomous vehicle tasks to accurately evaluate model performance and visualizations.

\section{Conclusion}

We conducted a comparative analysis of multiple model architectures and input configurations to predict pedestrian actions using raw images as input. Our results demonstrated that pure CNN- or Transformer-based networks outperformed hybrid architectures, which could be attributed to the former's pre-training on large corpus datasets, unlike the latter. Additionally, pre-training was found to be a critical factor in using static versus dynamic crop settings possibly because dynamic crop images were similar to those in the kinetics dataset. Furthermore, Transformer-based models' interpretability showed that the features used to predict pedestrian actions were more human-like in the dynamic mode than the static mode. In the dynamic mode, the model successfully localized the pedestrian's location and body keypoints. This could be attributed to the fact that the models were pre-trained to capture dynamic gestures of humans rather than a static environment. However, we also found it surprising that the static mode achieved reasonable quantitative results even though its qualitative results indicated the presence of non-relevant features. To make a fair comparison, it would be necessary to train a large model on datasets related to autonomous vehicles and then fine-tune it to perform the specific action prediction task.

\section*{Acknowledgement}
This work was carried out in the framework of the OpenLab ``Artificial Intelligence'' in the context of a partnership between INRIA institute and Stellantis company.

\end{document}